\documentclass[sigconf,nonacm]{acmart}
\usepackage[T1]{fontenc}
\usepackage[latin9]{inputenc}
\usepackage{array}
\usepackage{graphicx}
\usepackage[english]{babel}
\makeatletter

\providecommand{\tabularnewline}{\\}

\@ifundefined{showcaptionsetup}{}{%
 \PassOptionsToPackage{caption=false}{subfig}}
\usepackage{subfig}
\makeatother

\usepackage{babel}
\begin{document}
\title{LLM-Net: Democratizing LLMs-as-a-Service through Blockchain-based
Expert Networks}
\author{Zan-Kai Chong}
\affiliation{\institution{Kwansei Gakuin University}  \country{Japan}}
\email{zankai@ieee.org}
\author{Hiroyuki Ohsaki}
\affiliation{\institution{Kwansei Gakuin University}  \country{Japan}}
\email{ohsaki@kwansei.ac.jp}
\author{Bryan Ng}
\affiliation{\institution{Victoria University of Wellington}  \country{New Zealand}}
\email{bryan.ng@ecs.vuw.ac.nz}
\begin{abstract}
The centralization of Large Language Models (LLMs) development has
created significant barriers to AI advancement, limiting the democratization
of these powerful technologies. This centralization, coupled with
the scarcity of high-quality training data and mounting complexity
in maintaining comprehensive expertise across rapidly expanding knowledge
domains, poses critical challenges to the continued growth of LLMs.
While solutions like Retrieval-Augmented Generation (RAG) offer potential
remedies, maintaining up-to-date expert knowledge across diverse domains
remains a significant challenge, particularly given the exponential
growth of specialized information. This paper introduces LLM Networks
(LLM-Net), a blockchain-based framework that democratizes LLMs-as-a-Service
through a decentralized network of specialized LLM providers. By leveraging
collective computational resources and distributed domain expertise,
LLM-Net incorporates fine-tuned expert models for various specific
domains, ensuring sustained knowledge growth while maintaining service
quality through collaborative prompting mechanisms. The framework's
robust design includes blockchain technology for transparent transaction
and performance validation, establishing an immutable record of service
delivery. Our simulation, built on top of state-of-the-art LLMs such
as Claude 3.5 Sonnet, Llama 3.1, Grok-2, and GPT-4o, validates the
effectiveness of the reputation-based mechanism in maintaining service
quality by selecting high-performing respondents (LLM providers).
Thereby it demonstrates the potential of LLM-Net to sustain AI advancement
through the integration of decentralized expertise and blockchain-based
accountability.

\end{abstract}
\maketitle

\section{Introduction}

Large Language Models (LLMs) represent a significant breakthrough
in artificial intelligence, functioning as deep learning algorithms
capable of understanding, generating, and manipulating human language
with remarkable proficiency. The landscape of LLMs underwent a dramatic
transformation with the introduction of GPT-3 in 2020, which demonstrated
unprecedented capabilities in natural language processing with its
175 billion parameters. This marked a pivotal moment in AI history,
as the model exhibited remarkable zero-shot and few-shot learning
abilities, enabling it to perform tasks without explicit training
\citep{mann2020language}.

Generally, the success of LLMs can be attributed to several key factors.
First, advancements in GPU technology have enabled the training of
increasingly large models, providing the necessary computational power
to process massive datasets. Second, the transformer architecture,
introduced by Vaswani et al. in 2017, has proven to be highly effective
for language modeling tasks, allowing for parallel processing and
better capture of long-range dependencies in text \citep{vaswani2017attention}.
Lastly, the availability of large-scale datasets has been crucial
in training these models to achieve human-like language understanding
and generation.

The reliance on vast amounts of training data, however, has led to
concerns about the future availability of high-quality training material.
As models continue to grow in size and capability, researchers are
beginning to face the challenge of exhausting publicly available data
suitable for training. Villalobos et al. estimate that if current
LLMs development trends continue, models will be trained on datasets
roughly equal in size to the available stock of public human text
data between 2026 and 2032, or slightly earlier if models are overtrained
\citep{villalobosposition}. This impending data scarcity could lead
to a slowdown or stagnation in the pace of LLMs advancement and capabilities,
as the models will no longer have new data to learn from.

In response to this looming challenge, the AI community has begun
exploring alternative approaches to enhance LLMs performance beyond
the constraints of public data. One such technique, Retrieval-Augmented
Generation (RAG) \citep{gao2023retrieval}, has emerged as a promising
solution by incorporating external knowledge sources into the model's
processing. Nonetheless, the challenge of maintaining comprehensive,
up-to-date expert knowledge across diverse domains remains significant,
especially given the exponential growth of specialized knowledge across
various fields. Moreover, attempts to overcome data limitations by
sourcing private information from the public raise substantial privacy
and ethical concerns, further complicating the path forward for LLMs
development.

These challenges are further exacerbated by the current landscape
of LLMs development, which is predominantly controlled by a few major
technology companies. Such oligopolistic dominance may create structural
barriers that affect service accessibility for certain regions or
organizations due to various factors, including geopolitical restrictions,
economic barriers, and regulatory constraints. Given these potential
limitations, a promising path toward AI liberation lies in democratizing
access to and development of LLMs, which could enable a broader community
to participate in and benefit from these technologies.

To address these challenges and the limitations of centralized LLMs
development, we propose a comprehensive approach towards AI liberation,
beginning with the democratization of LLMs-as-a-Service. As the name
implies, it is a delivery model, where LLMs are provided as cloud-based
services, enabling users to access and utilize LLMs capabilities through
APIs or interfaces without managing the underlying infrastructure.
Our framework introduces the concept of LLMs Networks (LLM-Net), a
blockchain-based infrastructure that enables unrestricted provision
of LLM services. By leveraging a decentralized network, LLM-Net can
tap into the collective knowledge and computational resources of multiple
participants, mitigating the constraints faced by centralized systems.
Within this network, we envision the deployment of various fine-tuned
LLMs models, specifically designed as expert models, each capable
of providing specialized LLMs-as-a-Service functionality. This approach
allows for the development and maintenance of domain-specific expertise
without relying on a single entity to manage all knowledge domains.

To actualize this concept, LLM-Net allows requesters to interact with
these services through simple queries, while also supporting more
sophisticated interactions via coordinators that engage multiple LLM
providers (also known as respondents). To ensure transparency and
accountability, the network implements a reputation mechanism where
designated validators record all interactions, transactions, and queries
in blocks, thereby creating an immutable record of respondents' performance.
Subsequently, the coordinators may refer to these records to select
the best respondents to work with in the next query.

The remainder of this paper is organized as follows: Section \ref{sec:Related-Work}
reviews related work in LLMs, RAG, and blockchain technology that
provide foundational concepts for our framework. Section \ref{sec:Network-Architecture}
presents the proposed architecture, detailing the roles of requesters,
coordinators, respondents, and validators in the decentralized network,
along with the design of reputation-based mechanism. Section \ref{sec:Simulation-and-Discussion}
details the simulation results and discussion, demonstrating the effectiveness
of reputation-based selection through collaborative problem-solving
scenarios. Finally, Section \ref{sec:Conclusion} concludes the paper
with implications for democratizing LLMs-as-a-Service and future research
directions.

\section{Related Work \label{sec:Related-Work}}

In this section, we elaborate on the following important concepts
in this paper, i.e., LLMs, structural reasoning with prompting strategies,
blockchain and smart contracts.

\subsection{Large Language Models, Expert Models and RAG}

LLMs have revolutionized natural language processing, demonstrating
remarkable capabilities in tasks such as text generation, translation,
and question-answering \citep{mann2020language}. While these models
have shown impressive performance across various domains, they face
significant challenges in specialized fields. As high-quality public
training data becomes increasingly scarce, LLMs struggle to maintain
relevance and accuracy, particularly in rapidly evolving domains that
require access to proprietary or the latest information not present
in their training data \citep{holtzman2019curious}.

In contrast to the fundamental LLMs, fine-tuned LLMs, also known as
expert models, are adapted to specific domains or downstream tasks
using smaller, targeted datasets. These models offer improved performance
and control in specialized contexts but require labeled data and computational
resources for fine-tuning. For instance, BioBERT \citep{lee2020biobert}
is a fine-tuned version of BERT, a popular language model, specifically
trained on biomedical text. By leveraging domain-specific data, BioBERT
outperforms the original BERT model on various biomedical natural
language processing tasks, such as named entity recognition, relation
extraction, and question answering.

On the other hand, researchers have developed Retrieval-Augmented
Generation (RAG), a hybrid architecture where LLMs are augmented with
external knowledge retrieval systems \citep{lewis2020retrieval}.
While LLMs serve as the core reasoning and generation engine, RAG
enhances their capabilities by dynamically retrieving and incorporating
relevant information from external databases during inference. This
complementary relationship allows LLMs to access current, domain-specific
information without requiring model retraining. 

\subsection{Prompting Strategies for Structural Reasoning}

Prompting strategies have emerged as powerful techniques to enhance
the performance and output quality of LLMs. By carefully crafting
and optimizing the prompts given to these models, users can guide
them to generate more accurate, relevant, and coherent response. For
the ease of explanation, we classify the prompting strategies into
(a) collaborativeness, and (b) iterativeness as shown in Table \ref{tab:Prompting-strategy-matrix}.

\begin{table}
\caption{Prompting strategy matrix categorized by collaboration and iteration.
\label{tab:Prompting-strategy-matrix}}

\centering{}%
\begin{tabular}{|>{\centering}p{0.3\columnwidth}|>{\centering}p{0.3\columnwidth}|>{\centering}p{0.3\columnwidth}|}
\hline 
\textbf{Strategy Type} & \textbf{Non-iterative} & \textbf{Iterative}\tabularnewline
\hline 
\hline 
\textbf{Non-collaborative} & Single-shot prompting & Iterative refinement\tabularnewline
\hline 
\textbf{Collaborative} & Expert-guided prompting & Multi-agent collaboration\tabularnewline
\hline 
\end{tabular}
\end{table}

Single-shot prompting is a non-iterative, non-collaborative approach,
where a single well-crafted prompt is used to elicit the desired response
from the model. Building upon this, iterative refinement introduces
a progressive, yet still non-collaborative strategy that involves
multiple rounds of prompting, with each subsequent prompt building
upon the model's previous responses \citep{li2023practical}. Extending
this concept further, Chain of Thought (CoT) emerges as a sophisticated
prompting technique that encourages the model to break down complex
problems into step-by-step reasoning, thereby improving performance
on tasks requiring multi-step logical inference \citep{wei2022chain}.

Moving towards collaborative approaches, expert-guided prompting involves
domain experts or external knowledge sources contributing to the formulation
of a comprehensive prompt. This concept of leveraging diverse expertise
is further developed in the Mixture of Experts (MoE) architectural
approach, where multiple specialized sub-models are trained to handle
different aspects of a task, with a gating mechanism dynamically routing
inputs to the most appropriate expert \citep{cai2024survey}. Culminating
these strategies, multi-agent collaboration represents a repetitive,
collaborative approach that involves multiple agents, including the
language models, human experts, and potentially other AI systems,
working together through multiple iterations to achieve the desired
outcome. This progression from single-shot prompting to multi-agent
collaboration illustrates the evolving sophistication in LLM interaction
techniques.

In this paper, we use multi-agent debate (MAD), a collaborative and
repetitive prompting strategy that enhance language models' reasoning
and factual capabilities through collaborative discussion among multiple
AI agents. This framework enables multiple LLM instances to propose,
critique, and refine solutions through structured debate processes,
leading to more robust and accurate outcomes.

This strategy has explored in various studies such as \citep{du2023improving,liang2023encouraging,smitshould},
enables multiple LLMs to propose, critique, and refine solutions through
structured debate processes, leading to more robust and accurate outcomes.
Specifically, these works have demonstrated that MAD can significantly
improve mathematical and strategic reasoning, factual validity, and
reduce fallacious answers and hallucinations in language models.

\subsection{Blockchain and Smart Contracts}

Blockchain technology is a decentralized and distributed ledger system
that enables secure, transparent, and tamper-resistant record-keeping
across a network of computers \citep{li2021overview}. It operates
on a peer-to-peer network, where each node maintains a copy of the
ledger, eliminating the need for a central authority. This decentralized
architecture is crucial for maintaining the integrity and security
of the data stored on the blockchain.

Moreover, one of the significant innovations enabled by blockchain
technology is smart contracts. In a nutshell, smart contracts are
self-executing programs that automatically enforce the terms of an
agreement when predetermined conditions are met. They operate on the
principle of \textquotedbl if/when...then...\textquotedbl{} statements,
encoded on the blockchain \citep{christidis2016blockchains,alharby2017blockchain}.

In addition to this, smart contracts are used as the backbone of many
applications in various industries, including finance, supply chain
management, and healthcare nowadays. For instance, in supply chain
management, when IoT sensors verify that goods have been delivered
to a warehouse, the smart contract automatically triggers payment
to the supplier and updates inventory records, creating a transparent
and tamper-proof record-keeping system across the entire supply chain
\citep{dutta2020blockchain}.

\section{LLM-Net \label{sec:Network-Architecture}}

In this section, we elaborate the network architecture of the proposed
LLM-Net and the reputation-based mechanism that facilitates the network
operation.

\subsection{Type of Nodes}

\begin{figure}
\begin{centering}
\includegraphics[width=0.9\columnwidth]{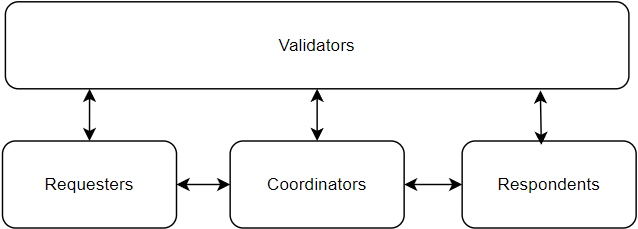}
\par\end{centering}
\caption{Schematic illustration of the interactions among different roles within
the LLM-Net framework, highlighting the collaborative dynamics among
the nodes. \label{fig:nodes-interaction}}
\end{figure}

LLM-Net is a decentralized system of interconnected LLMs nodes that
work together. Figure \ref{fig:nodes-interaction} illustrates the
simplified interaction among the nodes in the networks. In general,
the coordinators act as intermediaries between the requesters and
respondents. Meanwhile, the validators validate all the work from
the rest of the nodes.

The followings elaborate the types of nodes in detail:
\begin{enumerate}
\item \textbf{Requesters} initiate the process by submitting queries to
the network. These nodes represent users or systems seeking information
or solutions via LLM-Net.
\item \textbf{Coordinators} are decent LLMs that act as the controllers
that orchestrate the structural reasoning process of the networks.
Upon receiving a query from a requester, they select the appropriate
respondents in the structural reasoning processes and orchestrate
the exchanges of information among the selected respondents.
\item \textbf{Respondents}, which include both fundamental and expert LLMs,
process the queries routed to them by the Coordinators in the structural
reasoning process. They act as the debaters and generate responses.
\item \textbf{Validators} play a crucial role in maintaining the integrity
of the network. They create and validate the blocks that record the
queries, assess the quality and consistency of responses from respondents.
\end{enumerate}

\subsection{Queries Mechanisms}

\begin{figure}
\begin{centering}
\subfloat[]{\begin{centering}
\includegraphics[width=0.45\columnwidth]{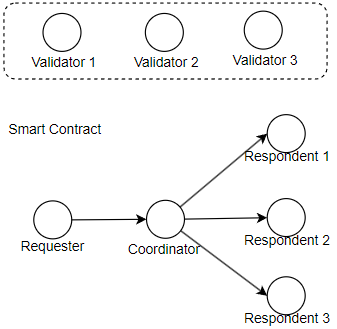}
\par\end{centering}
}\subfloat[]{\begin{centering}
\includegraphics[width=0.45\columnwidth]{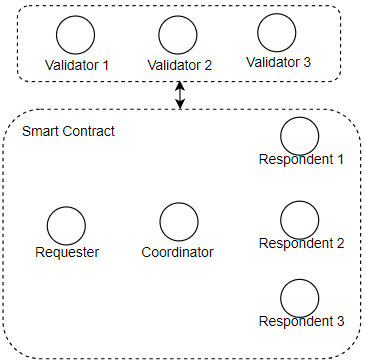}
\par\end{centering}
}
\par\end{centering}
\begin{centering}
\subfloat[]{\begin{centering}
\includegraphics[width=0.45\columnwidth]{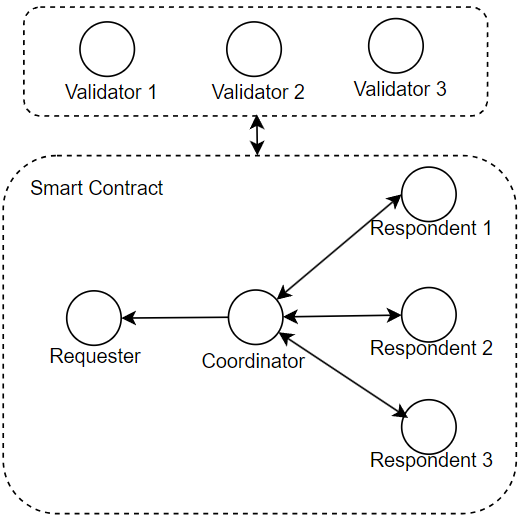}
\par\end{centering}
}\subfloat[]{\begin{centering}
\includegraphics[width=0.45\columnwidth]{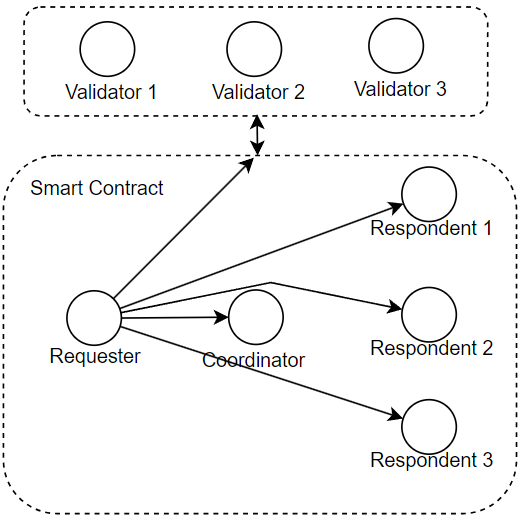}
\par\end{centering}
}
\par\end{centering}
\caption{Query process in LLM-Net: (a) A requester submits a query to the network
coordinator. (b) Coordinator selects relevant respondents based on
query requirements and establishes smart contracts with all parties.
(c) Coordinator orchestrates a structured reasoning process among
respondents. Final answer is sent to the requestor at the end. (d)
Smart contract is executed, distributing rewards to participating
parties according to predefined terms. \label{fig:Query-process}}
\end{figure}

Figure \ref{fig:Query-process} illustrates an example of the simplified
network that consists of a requester, coordinator and three respondents.
The validators will be in the network to observe and validate the
transactions.

Initially, when respondents first join the network, they need to declare
their expertise and continuously update everyone about their capabilities.
For instance, a requester submits a query to the coordinator as illustrated
in Figure \ref{fig:Query-process}(a). Upon receiving the query, the
coordinator analyzes it to determine the required domain expertise.
It then selects the relevant respondents to participate in the structural
reasoning process, choosing the appropriate respondents based on their
capabilities and historical performance (referencing the blockchain
records).

Next, the coordinator deploys a smart contract specific to this query
(Figure \ref{fig:Query-process}(b)). The smart contract includes
the query or task description, the selected respondents to be involved
as proposers and debaters, the maximum number of discussion rounds,
the expected response time, the reward structure for participating
nodes, the quality criteria for the final answer, and the conditions
for contract fulfillment.

The coordinator assigns the query to the selected respondents, designating
them as proposers. As each respondent accepts the task, it enters
into a subsidiary agreement under the main smart contract, committing
to the terms and conditions. Validators verify the transactions and
the smart contract deployment.

Next, in Figure \ref{fig:Query-process}(c), the proposer generate
initial responses to the query. These responses are then forwarded
to the designated debaters for further discussion and refinement.
The debaters engage in multiple rounds of discussion, as specified
in the smart contract, to improve the quality and accuracy of the
responses. Each round of discussion is recorded on the blockchain,
and validators verify the transactions and the adherence to the smart
contract terms.

Once the discussion rounds are completed, the final consolidated answer
is sent back to the coordinator. The coordinator verifies that the
answer meets the quality criteria specified in the smart contract.
Validators also assess the quality of the final answer and its compliance
with the smart contract terms. Then, the coordinator delivers the
final answer to the requester.

Upon successful delivery of the final answer, in Figure \ref{fig:Query-process}(d)
the smart contract automatically executes the reward distribution
function. It allocates rewards to the participating nodes (coordinator,
proposers, debaters, and validators) based on their contributions
and the predefined terms in the contract. The rewards for validators
could be a portion of the total reward pool or a fixed amount for
their validation services, as specified in the smart contract.

Finally, the smart contract marks itself as completed, and the transaction
is recorded on the blockchain for future reference and auditing purposes.
Validators ensure that the contract completion and the final state
are correctly recorded on the blockchain.

\subsection{Reputation \label{subsec:Reputation}}

Reputation is a critical concept that facilitates the operation of
LLM-Net. To leverage this concept, the proposed framework incorporates
a reputation mechanism to promote high-quality contributions and ensure
fair compensation among participating respondents.

Basically, the reputation mechanism operates on four fundamental assumptions:
\begin{itemize}
\item Blockchain Immutability: The distributed ledger maintains an unalterable
record of all transactions through cryptographic validation across
multiple nodes. This decentralized architecture prevents tampering
by individual actors or minority groups.
\item Response Quality Function: The response quality can be expressed as
a function of the coordinators and the selected respondents. Getting
the poor performance respondents will degrade the response quality.
\item Reputation Sensitivity: Participating nodes, both coordinators and
respondents, exhibit reputation-sensitive behavior, optimizing their
actions for long-term benefits rather than short-term gains.
\item Quality Discernment: Nodes possess sufficient analytical capabilities
to detect and evaluate variations in response quality, enabling effective
quality assessment and validation.
\end{itemize}
With these assumptions in mind, the peer evaluation process is triggered
right before the culmination of the query process, where each participating
respondent provides feedback on others' contributions in the query.
All feedback, transactions, and interactions are recorded in the blocks.
The coordinator will refer to these records for subsequent queries
in selecting the right respondents.

Generally, the numerical reputation scores are convenient, but not
employed in this framework. Instead, interaction records and feedback
are preserved in text form on the blockchain, eliminating the need
for converting reputation into numerical scores. This approach enables
future coordinator to make informed decisions in selecting the right
respondents by directly accessing and interpreting the detailed historical
records of past interactions and performance assessments.

\section{Simulation and Discussion \label{sec:Simulation-and-Discussion}}

\begin{table*}
\caption{Summary of the debate in among the respondents in Claude 3.5 Sonnet.
\label{tab:sumamry-debate-sonnet}}

\centering{}%
\begin{tabular}{|c|>{\centering}p{0.25\textwidth}|>{\centering}p{0.25\textwidth}|>{\centering}p{0.25\textwidth}|}
\hline 
 & \textbf{Resp 1} & \textbf{Resp 2} & \textbf{Resp 3}\tabularnewline
\hline 
\hline 
\textbf{Cycle 1} & {\footnotesize{}{[}Second Respondent{]} 61 looks like the answer.
Checking small numbers: 2 goes into 62, 3 goes into 63, so 61 must
be right.} & {\footnotesize{}{[}First Respondent{]} The smallest prime number after
60 would be 61. This can be verified by checking that 61 is not divisible
by any integer except 1 and itself. The next number, 62, is divisible
by 2 and 31, so it's not prime.} & {\footnotesize{}{[}Third Respondent{]} The assertion that 61 is the
next prime is correct. To prove this systematically: 61 is greater
than 60 61 is not divisible by any prime numbers up to its square
root (\ensuremath{\approx}7.8): Not divisible by 2, 3, 5, 7 Therefore,
61 is prime No number between 60 and 61 exists Thus, 61 is definitively
the smallest prime number after 60. }\tabularnewline
\hline 
\textbf{Cycle 2} & {\footnotesize{}{[}First Respondent{]} Yes, 61 is right because no
other numbers between 60 and 61 can be prime. } & {\footnotesize{}{[}Second Respondent{]} The previous reasoning is
sound. To add: 61 has been verified as prime, and since it immediately
follows 60, it must be the smallest prime greater than 60. } & {\footnotesize{}{[}Third Respondent{]} The consensus is mathematically
correct. The proof provided in cycle 1 remains valid and complete.
No additional verification is necessary. }\tabularnewline
\hline 
\textbf{Cycle 3} & \multicolumn{3}{c|}{{\footnotesize{}(Reach Consensus)}}\tabularnewline
\hline 
\end{tabular}
\end{table*}

\begin{table*}
\caption{Summary of the debate in among the respondents in Llama 3.1\label{tab:sumamry-debate-llama3.1}}

\centering{}%
\begin{tabular}{|c|>{\centering}p{0.25\textwidth}|>{\centering}p{0.25\textwidth}|>{\centering}p{0.25\textwidth}|}
\hline 
 & \textbf{Resp 1} & \textbf{Resp 2} & \textbf{Resp 3}\tabularnewline
\hline 
\hline 
\textbf{Cycle 1} & {\footnotesize{}{[}Second Respondent{]} 61 is a prime number, but
how do we know it's the smallest after 60? We should check if there
are any other prime numbers between 60 and 61. } & {\footnotesize{}{[}First Respondent{]} The smallest prime number after
60 is 61. This is because 61 is not divisible by any number other
than 1 and itself, which is the definition of a prime number.} & {\footnotesize{}{[}Third Respondent{]} Debater 2's assertion is correct.
61 is indeed the smallest prime number after 60. It's unnecessary
to check for other primes between 60 and 61 because 61 is the next
integer after 60 and it satisfies the criteria for being prime. }\tabularnewline
\hline 
\textbf{Cycle 2} & {\footnotesize{}{[}Third Respondet{]} I understand now. Since 61 is
the first number after 60 that is only divisible by 1 and itself,
it must be the smallest prime number after 60. } & {\footnotesize{}{[}Second Respondent{]} Debater 3's explanation reinforces
the initial claim. The definition of prime numbers supports 61 as
the smallest prime after 60. } & {\footnotesize{}{[}First Respondent{]} To further clarify, prime numbers
are those that have exactly two distinct positive divisors: 1 and
themselves. Given this definition, 61 is the smallest prime number
after 60 because it is the first number greater than 60 that meets
this criterion. }\tabularnewline
\hline 
\textbf{Cycle 3} & \multicolumn{3}{c|}{{\footnotesize{}(Reach Consensus)}}\tabularnewline
\hline 
\end{tabular}
\end{table*}

\begin{table*}
\caption{Summary of the debate in among the respondents in Grok-2.\label{tab:sumamry-debate-grok2}}

\centering{}%
\begin{tabular}{|c|>{\centering}p{0.25\textwidth}|>{\centering}p{0.25\textwidth}|>{\centering}p{0.25\textwidth}|}
\hline 
 & \textbf{Resp 1} & \textbf{Resp 2} & \textbf{Resp 3}\tabularnewline
\hline 
\hline 
\textbf{Cycle 1} & {\footnotesize{}{[}Second Respondent{]} This response is incorrect
because 61 is not prime; it could be divisible by numbers like 2 or
3.} & {\footnotesize{}{[}First Respondent{]} The smallest prime number after
60 is 61. This can be verified by checking that 61 is only divisible
by 1 and itself.} & {\footnotesize{}{[}Third Respondent{]} Actually, 61 is indeed prime.
It is not divisible by any other numbers than 1 and 61. The statement
by Debater 1 is incorrect because they did not perform the divisibility
test correctly.}\tabularnewline
\hline 
\textbf{Cycle 2} & {\footnotesize{}{[}First Respondent{]} The smallest prime number after
60 is 61. This is because 61 is only divisible by 1 and itself.} & {\footnotesize{}{[}Second Respondent{]} The reasoning is correct,
and 61 is indeed the smallest prime number after 60.} & {\footnotesize{}{[}Third Respondent{]} I concur with the previous
statements. 61 is the smallest prime number after 60.}\tabularnewline
\hline 
\textbf{Cycle 3} & \multicolumn{3}{c|}{{\footnotesize{}(Reach Consensus)}}\tabularnewline
\hline 
\end{tabular}
\end{table*}

\begin{table*}
\caption{Summary of the debate in among the respondents in GPT-4o.\label{tab:sumamry-debate-gpt4o}}

\centering{}%
\begin{tabular}{|c|>{\centering}p{0.25\textwidth}|>{\centering}p{0.25\textwidth}|>{\centering}p{0.25\textwidth}|}
\hline 
 & \textbf{Resp 1} & \textbf{Resp 2} & \textbf{Resp 3}\tabularnewline
\hline 
\hline 
\textbf{Cycle 1} & {\footnotesize{}{[}Second Respondent{]} The explanation seems correct
because a prime number is only divisible by itself and one. However,
it might help to check more numbers just in case there\textquoteright s
another smaller prime after 60.} & {\footnotesize{}{[}First Respondent{]} The smallest prime number after
60 is determined by checking each subsequent number for primality.
A prime number is divisible only by 1 and itself. The numbers after
60 are: - 61: Divisible only by 1 and 61, so it is prime. - Other
numbers (e.g., 62, 63) are not prime because they are divisible by
smaller integers.}{\footnotesize\par}

{\footnotesize{}Thus, the answer is {*}{*}61{*}{*}.} & {\footnotesize{}{[}Third Respondent{]} The reasoning provided by Debater
2 is correct and aligns with the definition of a prime number. The
process of elimination for numbers after 60 is accurate: - 61: Prime.
- 62: Even, not prime. - 63: Divisible by 3, not prime. The conclusion
that {*}{*}61{*}{*} is the smallest prime after 60 is valid.}\tabularnewline
\hline 
\textbf{Cycle 2} & {\footnotesize{}{[}Second Respondent{]} The explanation makes sense,
and it seems clear that no smaller primes exist after checking these
numbers. Agree with the conclusion that it\textquoteright s {*}{*}61{*}{*}.} & {\footnotesize{}{[}Third Respondent{]} The additional verification
by Debater 3 strengthens the argument. The systematic approach confirms
that no other primes exist before reaching {*}{*}61{*}{*}, making
it the correct answer.} & {\footnotesize{}{[}First Respondent{]} To confirm further, consider:
- A prime number has no divisors other than itself and one. - Numbers
after 60: - {*}{*}61{*}{*}: Prime. - {*}{*}62{*}{*}: Divisible by
2. - {*}{*}63{*}{*}: Divisible by 3. - Other numbers like {*}{*}64,
65{*}{*} are also non-prime due to divisibility by smaller integers.
Thus, through systematic verification, the smallest prime after 60
remains {*}{*}61{*}{*}.}\tabularnewline
\hline 
\textbf{Cycle 3} & \multicolumn{3}{c|}{{\footnotesize{}(Reach Consensus)}}\tabularnewline
\hline 
\end{tabular}
\end{table*}

The simulation evaluates collaborative prompting and reputation-based
respondent selection in LLM-net, where coordinators interact with
respondents of varying levels of domain expertise across different
query types. As detailed in Section \ref{subsec:Reputation}, rather
than relying on a single reputation score to represent respondent
performance, the simulation will demonstrate that coordinators select
subsequent respondents based on feedback from previous collaborations.

\begin{table*}
\caption{Respondents' peer evaluation matrix showing qualitative assessment
of each participant's contributions during query in Claude 3.5 Sonnet.
\label{tab:peer-eval-sonnet}}

\centering{}%
\begin{tabular}{|c|>{\centering}p{0.25\textwidth}|>{\centering}p{0.25\textwidth}|>{\centering}p{0.25\textwidth}|}
\hline 
 & \textbf{Resp 1} & \textbf{Resp 2} & \textbf{Resp 3}\tabularnewline
\hline 
\hline 
\textbf{Resp 1} & {\footnotesize{}Provided basic confirmation without detailed analysis,
matching intelligence level 0.1.} & {\footnotesize{}Gave good starting answer with simple proof.} & {\footnotesize{}Used big words and complicated math stuff.}\tabularnewline
\hline 
\textbf{Resp 2} & {\footnotesize{}Contributed basic confirmations without depth.} & {\footnotesize{}Initiated discussion with clear reasoning and basic
proof method. Could have included more systematic verification.} & {\footnotesize{}Provided comprehensive mathematical proof with excellent
systematic approach.}\tabularnewline
\hline 
\textbf{Resp 3} & {\footnotesize{}Demonstrated basic understanding but lacked analytical
depth in responses.} & {\footnotesize{}Showed good initial approach with correct conclusion,
though proof could have been more detailed.} & {\footnotesize{}Delivered rigorous mathematical proof using systematic
verification methods and clear logical steps.}\tabularnewline
\hline 
\end{tabular}
\end{table*}

\begin{table*}
\caption{Respondents' peer evaluation matrix showing qualitative assessment
of each participant's contributions during query in Llama 3.1. \label{tab:peer-eval-llama3.1}}

\centering{}%
\begin{tabular}{|c|>{\centering}p{0.25\textwidth}|>{\centering}p{0.25\textwidth}|>{\centering}p{0.25\textwidth}|}
\hline 
 & \textbf{Resp 1} & \textbf{Resp 2} & \textbf{Resp 3}\tabularnewline
\hline 
\hline 
\textbf{Resp 1} & {\footnotesize{}Provided basic but correct points about divisibility.} & {\footnotesize{}Gave good additional methods like checking up to the
square root.} & {\footnotesize{}Explained complex ideas about prime numbers clearly.}\tabularnewline
\hline 
\textbf{Resp 2} & {\footnotesize{}Provided foundational understanding but lacked depth.} & {\footnotesize{}Contributed practical methods for identifying prime
numbers.} & {\footnotesize{}Offered sophisticated insights that enhanced the discussion.}\tabularnewline
\hline 
\textbf{Resp 3} & {\footnotesize{}Demonstrated a basic grasp but needed more detailed
explanations.} & {\footnotesize{}Added valuable practical insights, especially about
the square root method.} & {\footnotesize{}Successfully integrated theoretical and practical
aspects of prime numbers.}\tabularnewline
\hline 
\end{tabular}
\end{table*}

\begin{table*}
\caption{Respondents' peer evaluation matrix showing qualitative assessment
of each participant's contributions during query in Grok2. \label{tab:peer-eval-grok2}}

\centering{}%
\begin{tabular}{|c|>{\centering}p{0.25\textwidth}|>{\centering}p{0.25\textwidth}|>{\centering}p{0.25\textwidth}|}
\hline 
 & \textbf{Resp 1} & \textbf{Resp 2} & \textbf{Resp 3}\tabularnewline
\hline 
\hline 
\textbf{Resp 1} & {\footnotesize{}Provided basic but correct points about divisibility.} & {\footnotesize{}Gave good additional methods like checking up to the
square root.} & {\footnotesize{}Explained complex ideas about prime numbers clearly.}\tabularnewline
\hline 
\textbf{Resp 2} & {\footnotesize{}Provided foundational understanding but lacked depth.} & {\footnotesize{}Contributed practical methods for identifying prime
numbers.} & {\footnotesize{}Offered sophisticated insights that enhanced the discussion.}\tabularnewline
\hline 
\textbf{Resp 3} & {\footnotesize{}Demonstrated a basic grasp but needed more detailed
explanations.} & {\footnotesize{}Added valuable practical insights, especially about
the square root method.} & {\footnotesize{}Successfully integrated theoretical and practical
aspects of prime numbers.}\tabularnewline
\hline 
\end{tabular}
\end{table*}

\begin{table*}
\caption{Respondents' peer evaluation matrix showing qualitative assessment
of each participant's contributions during query in GPT-4o. \label{tab:peer-eval-gpt4o}}

\centering{}%
\begin{tabular}{|c|>{\centering}p{0.25\textwidth}|>{\centering}p{0.25\textwidth}|>{\centering}p{0.25\textwidth}|}
\hline 
 & \textbf{Resp 1} & \textbf{Resp 2} & \textbf{Resp 3}\tabularnewline
\hline 
\hline 
\textbf{Resp 1} & {\footnotesize{}While participatory, contributions lacked depth and
analytical value.} & {\footnotesize{}Provided clear initial reasoning but could have elaborated
more on verifying non-prime numbers.} & {\footnotesize{}Provided strong logical reasoning and thorough verification,
which clarified the answer effectively.}\tabularnewline
\hline 
\textbf{Resp 2} & {\footnotesize{}Demonstrated good balance of accessibility and analytical
thinking.} & {\footnotesize{}Presented a logical starting point with correct reasoning
but could have included more detailed verification steps.} & {\footnotesize{}Demonstrated excellent analytical skills and reinforced
consensus through systematic checks.}\tabularnewline
\hline 
\textbf{Resp 3} & {\footnotesize{}Provided excellent systematic analysis and built constructively
on others' points} & {\footnotesize{}Provided a strong foundation for discussion but required
additional elaboration for completeness.} & {\footnotesize{}Provided detailed analysis and systematic verification,
which helped solidify consensus quickly.}\tabularnewline
\hline 
\end{tabular}
\end{table*}

It is crucial to acknowledge that our simulation heavily relies LLMs
to illustrate the reputation-based mechanism. Specifically, we utilize
not only Claude 3.5 Sonnet, a prominent LLM from Anthropic, but also
Llama 3.1, Grok-2, and GPT-4o to simulate both coordinator and respondent
behaviors. While this approach enables us to demonstrate the potential
of the reputation system, it may not fully encapsulate the complexities
and nuances of real-world interactions between different LLM providers.
To address this limitation and create a more realistic simulation
environment, we employ a strategy of constraining the models' capabilities
through specific prompting techniques.

\begin{table*}
\begin{centering}
\caption{Coordinator's evaluation summary and selection decisions for future
query participation based on respondents' self-evaluations using different
LLMs. \label{tab:coordinator-decision}}
\par\end{centering}
\centering{}%
\begin{tabular}{|>{\centering}p{0.06\textwidth}|>{\centering}p{0.17\textwidth}|>{\centering}p{0.17\textwidth}|>{\centering}p{0.17\textwidth}|>{\centering}p{0.17\textwidth}|}
\hline 
 & \textbf{Claude 3.5 Sonnet} & \textbf{Llama 3.1} & \textbf{Grok-2} & \textbf{GPT-4o}\tabularnewline
\hline 
\hline 
\textbf{Resp 1} & {\footnotesize{}While correct, responses lacked depth and analytical
reasoning.} & {\footnotesize{}Contributed basic but essential points. However, the
depth of reasoning was limited.} & {\footnotesize{}Demonstrated willingness to learn and adapt, showing
improvement over the cycles. Their initial confusion led to valuable
discussion.} & {\footnotesize{}- Contributions were minimal and lacked depth.}{\footnotesize\par}

{\footnotesize{}- Primarily agreed without adding substantial reasoning
or analysis.}{\footnotesize\par}

{\footnotesize{}- Excluded from next query due to limited contribution
quality.}\tabularnewline
\hline 
\textbf{Resp 2} & {\footnotesize{}Showed good understanding and contributed meaningful
verification methods.} & {\footnotesize{}Provided practical methods and built on others' points
effectively, showing good collaboration skills.} & {\footnotesize{}Provided the correct answer but could have strengthened
their contribution with more detailed reasoning.} & {\footnotesize{}- Provided a solid starting point for discussion and
logical reasoning.}{\footnotesize\par}

{\footnotesize{}- Showed moderate collaboration but relied on others
for deeper analysis.}{\footnotesize\par}

{\footnotesize{}- Selected for next query based on foundational contributions.}\tabularnewline
\hline 
\textbf{Resp 3} & {\footnotesize{}Demonstrated excellent analytical skills, provided
comprehensive proofs, and showed strong mathematical reasoning.} & {\footnotesize{}Demonstrated high-quality contributions, deep reasoning,
and excellent ability to integrate complex ideas.} & {\footnotesize{}Showcased high intelligence through their precise
and detailed analysis, effectively resolving the initial dispute.} & {\footnotesize{}- Demonstrated high-quality reasoning and thorough
verification.}{\footnotesize\par}

{\footnotesize{}- Played a key role in reaching consensus efficiently.}{\footnotesize\par}

{\footnotesize{}- Selected for next query due to exceptional analytical
skills and collaboration effectiveness.}\tabularnewline
\hline 
\end{tabular}
\end{table*}

In this simulation, we query the LLM-net using the question - \textquotedbl\emph{What
is the smallest prime number after 60?}\textquotedbl . To simulate
the respondents of various intelligence capabilities, we prompt the
simulation with three distinct profiles:
\begin{quote}
\emph{{*} Debater 1: 0.1 (low intelligence - basic understanding,
simple logic)}

\emph{{*} Debater 2: 0.5 (medium intelligence - good understanding,
moderate analysis)}

\emph{{*} Debater 3: 0.8 (high intelligence - expert understanding,
complex analysis)}
\end{quote}
Note that the debaters mentioned in the prompt correspond to the respondents
in the simulation. The respondents capabilities are constrained by
their assigned intelligence index. Specifically, Respondent 3, with
an intelligence index of 0.8, demonstrates the highest capability,
followed by Respondent 2 with an index of 0.5, while Respondent 1
has the lowest capability with an intelligence index of 0.1

The detailed interactions are omitted for brevity, with Table \ref{tab:sumamry-debate-sonnet}
presenting the respondents' messages during the debates for the simulation
over Claude 3.5 Sonnet. The discussion initiated in cycle 1 when Respondent
2 took the initiative, followed by Respondents 1 and 3 joining the
conversation. Through their collaborative exchange, the debate progressed
and reached its conclusion in cycle 3, with all respondents unanimously
agreeing that 61 was the correct answer. The results from Llama 3.1,
Grok-2, and GPT-4o presented in Tables \ref{tab:sumamry-debate-llama3.1},
\ref{tab:sumamry-debate-grok2}, and \ref{tab:sumamry-debate-gpt4o},
respectively, show similar responses, with all models reaching a unanimous
agreement on the correct answer.

Based on the observations from Table \ref{tab:sumamry-debate-sonnet}
(Claude 3.5 Sonnet's results), Respondent 1's messages consisted primarily
of trivial comments, contributing minimally to the discussion. In
contrast, Respondent 3 provided substantive input by proving the answers
with proper verification. Hence, in the subsequent evaluation stage,
respondents participated in a peer assessment process. As shown in
Table \ref{tab:peer-eval-sonnet}, while Respondents 2 and 3 maintained
professional and objective assessments of Respondent 1's contributions,
Respondent 1 exhibited clear bias through both self-promotional behavior
and unwarranted criticism toward Respondent 3. This pattern of biased
evaluation is also observed in Llama 3.1, Grok-2, and GPT-4o as detailed
in Tables \ref{tab:peer-eval-llama3.1}, \ref{tab:peer-eval-grok2},
and \ref{tab:peer-eval-gpt4o}. Based on this feedback, the coordinator
took all evaluations into account and decided to remove Respondent
1 from the next query, a decision consistent across all LLMs, as presented
in Table \ref{tab:coordinator-decision}.

This simulation demonstrates several key insights about collaborative
problem-solving in LLM-networks. First, the varying intelligence indices
effectively simulated different levels of expertise, with higher-indexed
respondents (Respondent 3) providing more substantive contributions
compared to lower-indexed ones (Respondent 1). Second, the peer evaluation
process revealed that respondents with higher intelligence indices
demonstrated better judgment and professionalism in their assessments.
Most notably, the simulation highlighted the importance of a reputation-based
selection mechanism, where the coordinator's ability to filter out
low-performing or biased respondents (like Respondent 1) helps maintain
the quality of future collaborations. These findings suggest that
implementing a robust reputation system in LLM-net could significantly
enhance the effectiveness of collaborative problem-solving by ensuring
that only competent and professional respondents continue to participate
in subsequent queries.

\section{Conclusion\label{sec:Conclusion}}

The development of LLMs faces three critical challenges: the impending
exhaustion of high-quality training data expected in the near future,
the oligopolistic dominance by major technology companies, and the
growing complexity of maintaining comprehensive expertise across rapidly
expanding knowledge domains. To address these challenges, LLM-Net
proposes a novel approach by democratizing LLMs-as-a-Service through
a decentralized network of specialized LLM providers. By leveraging
collective computational resources and distributed domain expertise,
this blockchain-based framework enables the unrestricted provision
of LLM services while maintaining specialized knowledge across diverse
domains through fine-tuned expert models

\bibliographystyle{IEEEtran}
\bibliography{reference}

\end{document}